\documentclass[runningheads]{llncs}

\usepackage[mobile]{eccv}

\usepackage{eccvabbrv}

\usepackage{graphicx}
\usepackage{booktabs}
\usepackage{capt-of}

\usepackage{caption}
\usepackage{subcaption}
\usepackage{float}
\usepackage{algorithm}%
\usepackage{algorithmicx}%
\usepackage{algpseudocode}%
\usepackage{paralist}

\usepackage{pdflscape}
\usepackage{multirow}
\usepackage{amssymb}
\usepackage[table]{xcolor}
\usepackage{xcolor}
\usepackage{comment}

\usepackage[protrusion=true,expansion=true]{microtype}
\usepackage[pagebackref]{hyperref}
\hypersetup{
  pdftitle={EsaacSim: A Multimodal Event Camera Add-on for NVIDIA Isaac Sim},
  pdfauthor={Anonymized},
  pdfsubject={Robotics (cs.RO); Computer Vision and Pattern Recognition (cs.CV)},%
  pdfkeywords={Event camera simulation, NVIDIA Isaac Sim, Neuromorphic vision, Robotics simulation, ROS~2},%
}

\usepackage{tikz}
\usetikzlibrary{arrows.meta,positioning,fit,calc,backgrounds}

\newcommand{\addon}{\textsc{EsaacSim}\xspace}

\usepackage[accsupp]{axessibility}  

\usepackage{hyperref}

\usepackage{orcidlink}

\begin{document}

\title{EsaacSim: A Multimodal Event Camera Add-on for NVIDIA Isaac Sim}

\titlerunning{EsaacSim}

\author{
Ignacio Bugueno-Cordova\inst{1,2}\orcidlink{0000-0002-0133-0330} \and
Malte Kuhlmann\inst{1}\orcidlink{0009-0008-3222-2827} \and
Nicolás Navarro-Guerrero\inst{1}\orcidlink{0000-0003-1164-5579} \and
Miguel Campusano\inst{3}\orcidlink{0000-0002-7894-6635} \and
Rodrigo Verschae\inst{4}\orcidlink{0000-0002-1661-3309}
}

\authorrunning{I. Bugueno-Cordova et al.}

\institute{
Leibniz Universität Hannover, L3S Research Center, Hanover, Germany\\
\and
Ekumen, Buenos Aires, Argentina\\
\and
University of Southern Denmark, Odense, Denmark\\
\and
The Iniciativa de Datos e Inteligencia Artificial,
University of Chile, Santiago, Chile\\
\email{i.bugueno@ieee.org}, 
\email{malte.kuhlmann@l3s.de}, 
\email{nicolas.navarro.guerrero@gmail.com}, 
\email{mica@mmmi.sdu.dk}, 
\email{rodrigo@verschae.org}
}

\maketitle

\begin{abstract}
    Event-based vision is becoming an increasingly important sensing paradigm for robotics, yet its adoption remains limited by sensor availability and the lack of integrated simulation tools for modern robotics platforms. This paper presents \addon, a multimodal event camera add-on for NVIDIA Isaac Sim that enables online simulation of configurable event cameras with grayscale and Bayer RGGB event generation. The framework supports multiple event camera resolutions and provides synchronized RGB, APS, event, depth, and IMU outputs through native ROS~2 interfaces. A motion-guided frame-gap synthesis strategy further increases the effective temporal resolution while preserving compatibility with the Isaac Sim rendering pipeline. Experimental evaluation demonstrates synchronized multimodal simulation across representative robotic scenes and efficient online performance over five event camera resolutions at effective event rates from 240 to 960~Hz. Event stream generation requires 6.98--27.28~ms for grayscale events and 7.58--29.16~ms for Bayer RGGB events while using less than 400~MB of additional GPU memory on an NVIDIA RTX~4060 GPU. These results show that \addon enables supports online multimodal event-camera simulation for robotics research and synthetic data generation. We release an early version of the simulator and report its current architecture and performance.
\end{abstract}

\section{Introduction}
\label{sec:intro}
Event-based cameras are attracting increasing attention in robotics due to their low latency, high dynamic range, and high temporal resolution \cite{gallego2019-survey,chakravarthi2024recent,cazzato2024}. These characteristics make them well suited for perception tasks in highly dynamic environments, including autonomous navigation, aerial robotics, robotic manipulation, and visual tracking \cite{wang2026,zafar2026,shariff2024,tenzin2024,hong2025}. Recent studies have demonstrated their advantages over conventional frame-based cameras under challenging motion and illumination conditions~\cite{durr2026ace}.
Despite these advantages, the development of event-based robotic systems remains constrained by the limited availability of sensors and large-scale event datasets~\cite{zafar2026}. Event camera simulators have therefore become an important alternative for synthetic data generation and algorithm development \cite{hageman2024survey}. Representative examples include ESIM \cite{rebecq2018-esim}, V2E \cite{delbruck2020-v2e}, and more recent simulation frameworks \cite{visapp25,li2025eventtracerfastpathtracingbased,greene2025pytorch}. However, few support online event generation, and most operate as standalone tools or dataset-generation pipelines~\cite{hageman2024survey}.

At the same time, robotics research increasingly relies on simulation-centric development workflows. NVIDIA Isaac Sim has emerged as a prominent robotics simulation platform by combining photorealistic RTX rendering, GPU-accelerated physics simulation, synthetic data generation, and integration with modern robot learning frameworks~\cite{gao2026nvidia,bonetto2026grade,jacinto2024,10610026}. Recently, EVIS~\cite{shi2026evis} introduced a physics-grounded event camera plugin for Isaac Sim, marking an important step toward native event camera simulation. While EVIS primarily focuses on synchronized RGB and event generation through a Python API, \addon targets robotics-oriented simulation by providing configurable sensor models, support for multiple event camera resolutions, native Bayer RGGB event generation, synchronized RGB, APS, event, depth, and IMU outputs, and seamless ROS~2 integration.

Motivated by these requirements, this paper presents \addon, a multimodal event camera add-on for NVIDIA Isaac Sim capable of generating online event streams from grayscale and Bayer RGGB image sequences. The proposed framework provides configurable event sensor simulation together with synchronized RGB, APS, event, depth, and IMU outputs through native ROS~2 interfaces. Furthermore, it supports multiple event camera resolutions through a unified architecture, enabling the simulation of a broad range of existing event sensor configurations.

The main contributions of this work are:
\begin{itemize}
    \item A multimodal online event camera add-on for NVIDIA Isaac Sim supporting multiple event camera resolutions, grayscale event generation, and native Bayer RGGB event generation together with synchronized RGB, Active Pixel Sensor frames (APS), depth, IMU, and event outputs through native ROS~2 interfaces.

    \item A configurable sensor modeling framework enabling the emulation of different event camera profiles through customizable thresholds, latency, refractory periods, and noise characteristics.

    \item An experimental evaluation including qualitative comparisons and performance benchmarking across multiple event camera resolutions and temporal interpolation factors.
\end{itemize}
The EsaacSim add-on will be publicly released.

\section{Related work}
The event camera simulation literature can be broadly divided into frame-based, physics-based, and robotics-oriented approaches. Table~\ref{tab:simulators_comparison} summarizes representative simulators and emulators according to their online capabilities, sensor modeling, robotics integration, supported simulation platforms, and generated data modalities.

\begin{table}[t]
\centering
\caption{Comparison of representative event camera simulators and emulators in terms of online operation, color event generation, configurable sensor modeling, physics-based event modeling, ROS~2 integration, availability within NVIDIA Isaac Sim, and generated data modalities. Output modalities are RGB images, APS frames, events (E), depth images (D), and IMU data. $^{*}$Physics-based event modeling for \addon\ is currently under development.}
\label{tab:simulators_comparison}
\begin{tabular}{@{}c|p{3.5cm}|c|c|c|c|c|c|p{1.8cm}@{}}
\toprule
Year &
Simulator &
Online &
\shortstack{Color\\Events} &
\shortstack{Sensor\\Model} &
\shortstack{Physics\\Model} &
\shortstack{ROS\\2} &
\shortstack{Isaac\\Sim} &
Outputs \\
\midrule

2012 & jAER Emulator \cite{katz2012}
& \checkmark &  &  &  &  &  & E \\

2016 & DVS Gazebo Plugin \cite{kaiser2016-dvsgazebo}
& \checkmark &  &  &  & \checkmark &  & E \\

2016 & pyDVS \cite{garcia2016-pydvs}
& \checkmark &  &  &  &  &  & E \\

2017 & PIX2NVS \cite{bi2017-pix2nvs}
&  &  &  &  &  &  & E \\

2018 & ESIM \cite{rebecq2018-esim}
&  &  & \checkmark & \checkmark &  &  & E \\

2020 & V2E \cite{delbruck2020-v2e}
&  &  & \checkmark &  &  &  & APS + E \\

2021 & AirSim Event Camera \cite{airsim-eventcamera}
& \checkmark &  &  &  & \checkmark &  & RGB + E \\

2021 & Joubert et al. \cite{joubert2021event}
&  &  & \checkmark & \checkmark &  &  & E \\

2024 & Color Event Emulator \cite{bugueno2024color}
& \checkmark & \checkmark & \checkmark &  & \checkmark &  & RGB + E \\

2024 & PECS \cite{han2024physical}
&  &  & \checkmark & \checkmark &  &  & E \\

2025 & EventTracer \cite{li2025eventtracerfastpathtracingbased}
&  &  & \checkmark & \checkmark &  &  & E \\

2025 & NeRF Simulator \cite{visapp25}
&  &  & \checkmark & \checkmark &  &  & E \\

2025 & SENPI \cite{greene2025pytorch}
&  &  & \checkmark & \checkmark &  &  & E \\

2026 & EVIS \cite{shi2026evis}
& \checkmark &  & \checkmark & \checkmark &  & \checkmark & RGB + E \\
\hline

2026 & \textbf{\addon\ (Ours)}
& \checkmark & \checkmark & \checkmark & \checkmark$^*$ & \checkmark & \checkmark &
RGB + APS + E + D + IMU \\
\bottomrule
\end{tabular}
\end{table}

\subsection{Frame-based Simulators and Emulators}
Early event emulation approaches demonstrated that conventional cameras could be used to approximate event-based sensing. Katz \textit{et al.} \cite{katz2012} integrated event emulation into the jAER framework using high-speed USB cameras, while pyDVS \cite{garcia2016-pydvs} and PIX2NVS \cite{bi2017-pix2nvs} enabled real-time and parameterized conversion from image sequences to event streams. More recently, V2E \cite{delbruck2020-v2e} and the real-time simulator of Ziegler \textit{et al.} \cite{ziegler2023} have become representative frame-based approaches for generating realistic event data from conventional imagery. Bugueno-Cordova \textit{et al.} \cite{bugueno2024color} extended this paradigm to color event cameras through a ROS-based online emulator for robot vision applications.

\subsection{Physics-based Simulators}
Physics-based simulators aim to model the event generation process more accurately by incorporating sensor characteristics, noise sources, optics, or direct interaction with 3D scenes. Representative examples include ESIM \cite{rebecq2018-esim}, the parameter-characterized simulator of Joubert \textit{et al.} \cite{joubert2021event}, differentiable event simulators for tracking and reconstruction \cite{nehvi2021}, and simulators designed for hardware prototyping and attention-based architectures \cite{pantho2022}. More recent works include PECS \cite{han2024physical}, which models the complete optical path, NeRF-based event synthesis from arbitrary viewpoints \cite{visapp25}, and EventTracer \cite{li2025eventtracerfastpathtracingbased}, which combines path tracing and event rendering to improve realism.

\subsection{Robotics-oriented Simulation Frameworks}
Several simulators have been developed specifically for robotics applications. The DVS Gazebo Plugin \cite{kaiser2016-dvsgazebo} integrates event sensing into Gazebo, while AirSim \cite{shah2017-airsim,airsim-eventcamera} provides event camera support within Unreal Engine. Similarly, Mueggler \textit{et al.} \cite{mueggler2017} and InteriorNet/ViSim \cite{li2018-interiornet} provide simulation environments and datasets for tasks such as visual odometry, SLAM, and autonomous navigation. Recently, EVIS \cite{shi2026evis} introduced a physics-grounded event camera plugin for NVIDIA Isaac Sim, representing the first event camera integration within this simulator. While EVIS demonstrates the feasibility of native event simulation in Isaac Sim through a physics-grounded sensor model, it primarily focuses on event generation itself.
As summarized in Table~\ref{tab:simulators_comparison}, existing simulators typically generate event streams alone or, in some cases, RGB images together with events. In contrast, \addon provides synchronized RGB images, event streams, depth images, and IMU measurements through native ROS~2 interfaces within Isaac Sim, facilitating the development of multimodal robotic perception pipelines.

\section{Event Generation Model}

\begin{figure}[!h]
    \centering
    \includegraphics[width=0.45\linewidth]{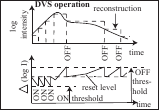}
    \caption{Temporal evolution of the logarithmic intensity signal \(L(t)\). Positive and negative events are generated whenever the accumulated logarithmic intensity variation reaches the corresponding contrast thresholds.}
    \label{fig:event_generation_model}
\end{figure}

Event cameras asynchronously generate events when changes in logarithmic image intensity exceed a contrast threshold. We adopt the standard event generation model illustrated in Figure~\ref{fig:event_generation_model}; comprehensive descriptions of event camera principles and hardware can be found in \cite{gallego2019-survey,shah2021review}.

For a pixel located at $\mathbf{u}_k=(x_k,y_k)^T$, an event is generated when
$
L(\mathbf{u}_k,t_k)-L(\mathbf{u}_k,t_k-\Delta t_k)\ge p_k C,
\label{eq:event_generation}
$
where $L=\log(I)$ denotes the logarithmic image intensity, $C$ is the contrast threshold, $p_k\in\{-1,+1\}$ is the event polarity, and $\Delta t_k$ is the elapsed time since the previous event generated at the same pixel. Positive and negative polarities correspond to increases and decreases in brightness, respectively. Consequently, the sensor output is represented as the asynchronous event stream
$
\mathcal{E}=\{(x_k,y_k,t_k,p_k)\}_{k=1}^{N}.
$

\section{\addon: An Event Camera Add-on for Isaac Sim}
\addon is an event camera add-on for NVIDIA Isaac Sim that enables online event simulation within photorealistic robotic environments. Rather than replacing the native RTX camera pipeline, \addon attaches to existing camera render products and converts their outputs into asynchronous grayscale or Bayer RGGB event streams. In contrast to EVIS~\cite{shi2026evis}, which primarily provides a physics-grounded event camera plugin for RGB-event simulation through a Python API, \addon is designed as a multimodal robotics add-on supporting configurable sensor models, multiple event camera resolutions, synchronized RGB, Activel Pixel Sensor (APS), event, depth, and IMU outputs, and native ROS~2 integration. 

\subsection{System Overview and Architecture}
\subsubsection{Functional Architecture}
\begin{figure}[!t]
    \centering
    \includegraphics[width=0.65\linewidth]{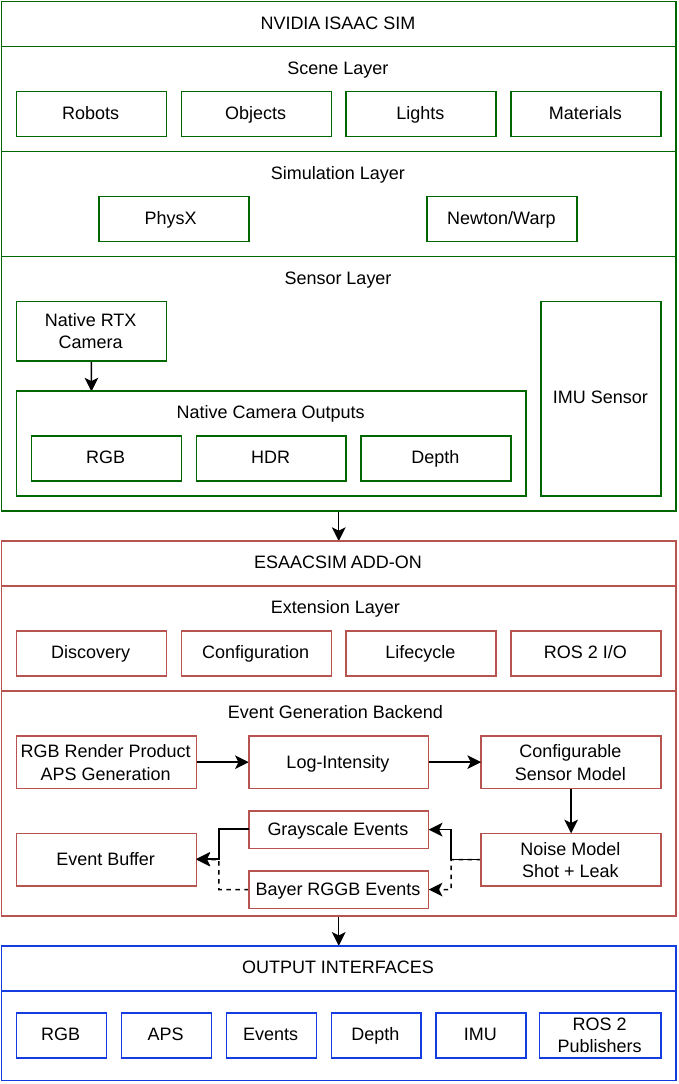}
    \caption{Layered architecture of \addon within NVIDIA Isaac Sim. Native Isaac Sim components (green) provide scene simulation and camera render products (RGB, HDR, and depth). \addon (red) acquires the RGB render products to generate APS frames and applies a configurable event camera model to produce either grayscale or Bayer RGGB event streams. RGB, APS, event, depth, and IMU data are synchronized and exposed through ROS 2 interfaces (blue).}
    \label{fig:esaac-sim-layer}
\end{figure}

\addon is organized into three logical layers, as illustrated in Fig.~\ref{fig:esaac-sim-layer}. The native Isaac Sim layer is responsible for scene representation, physics simulation, rendering, and camera image generation. Built on top of this infrastructure, the \addon extension layer discovers configured camera primitives, manages runtime configuration, acquires rendered image buffers through GPU-oriented ingestion paths, and provides visualization and ROS~2 communication. Finally, the event generation backend converts grayscale or Bayer image streams into asynchronous events using a frame-based log-intensity model with configurable sensor parameters before forwarding the generated event batches to the available output interfaces.
This layered organization decouples rendering, event generation, and data dissemination, allowing the event generation backend and output modules to evolve independently while preserving compatibility with the underlying Isaac Sim rendering pipeline.

Unlike EVIS~\cite{shi2026evis}, which focuses on a dedicated event camera plugin for RGB-event generation, \addon is organized as a multimodal sensing layer built on top of the native Isaac Sim sensor pipeline. This design enables synchronized RGB, APS, event, depth, and IMU outputs while preserving compatibility with existing Isaac Sim rendering, physics, and ROS~2 workflows.

\subsubsection{Runtime Deployment}
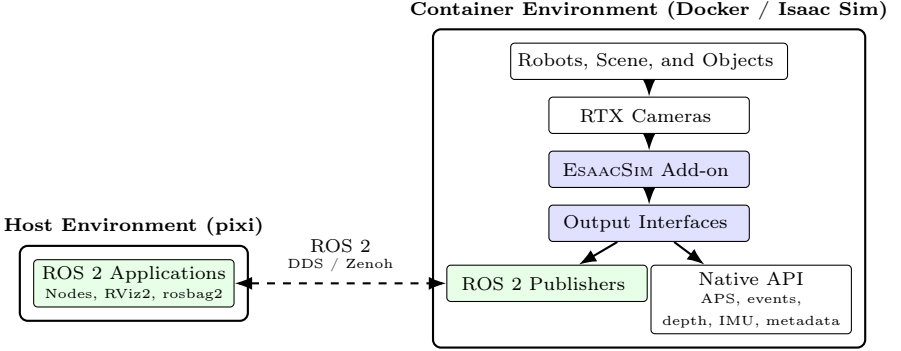
\begin{figure}[!t]
    \centering
    \begin{tikzpicture}[
        font=\scriptsize,
        >=Latex,
        block/.style={
            draw,
            rounded corners=1.5pt,
            minimum width=2.65cm,
            minimum height=0.48cm,
            align=center,
            inner sep=2.5pt,
            fill=white
        },
        addon/.style={
            block,
            fill=blue!12
        },
        ros/.style={
            block,
            fill=green!10
        },
        boundary/.style={
            draw,
            rounded corners=3pt,
            thick,
            inner sep=5pt
        },
        flow/.style={
            ->,
            thick
        },
        communication/.style={
            <->,
            thick,
            dashed
        },
        boundarytitle/.style={
            font=\scriptsize\bfseries
        }
    ]


    \node[block] (scene) {
        Robots, Scene, and Objects
    };

    \node[block, below=0.22cm of scene] (camera) {
        RTX Cameras
    };

    \node[addon, below=0.22cm of camera] (esaacsim) {
        \addon\ Add-on
    };

    \node[addon, below=0.22cm of esaacsim] (outputs) {
        Output Interfaces
    };

    \node[
        ros,
        below=0.30cm of outputs,
        xshift=-1.35cm
    ] (publisher) {
        ROS~2 Publishers
    };

    \node[
        block,
        below=0.30cm of outputs,
        xshift=1.35cm
    ] (native) {
        Native API\\[-1pt]
        \tiny APS, events,\\
        \tiny depth, IMU, metadata
    };

    \draw[flow] (scene) -- (camera);
    \draw[flow] (camera) -- (esaacsim);
    \draw[flow] (esaacsim) -- (outputs);
    \draw[flow] (outputs) -- (publisher);
    \draw[flow] (outputs) -- (native);

    \begin{scope}[on background layer]
        \node[
            boundary,
            fit=(scene)(camera)(esaacsim)(outputs)(publisher)(native),
            label={[boundarytitle]above:
                Container Environment (Docker / Isaac Sim)}
        ] (dockerbox) {};
    \end{scope}


    \node[
        ros,
        left=2.8cm of publisher
    ] (rosnodes) {
        ROS~2 Applications\\[-1pt]
        \tiny Nodes, RViz2, rosbag2
    };

    \begin{scope}[on background layer]
        \node[
            boundary,
            fit=(rosnodes),
            label={[boundarytitle]above:
                Host Environment (pixi)}
        ] (hostbox) {};
    \end{scope}


    \draw[communication]
        (rosnodes.east)
        --
        node[
            midway,
            above=0.08cm,
            align=center,
            fill=white,
            inner sep=2pt
        ] {
            ROS~2\\[-2pt]
            \tiny DDS / Zenoh
        }
        (publisher.west);

    \end{tikzpicture}
    \caption{
    Runtime deployment of \addon. The extension executes inside the
    Docker-based Isaac Sim environment and exposes generated data through
    ROS~2 publishers and native output interfaces. ROS~2 applications run
    in the host environment managed with \texttt{pixi}.
    }
    \label{fig:esaacsim-runtime-deployment}
\end{figure}

Figure~\ref{fig:esaacsim-runtime-deployment} illustrates the runtime deployment of \addon. The extension is loaded as a native Isaac Sim Kit add-on and executes inside the Docker-based simulation environment, where it receives rendered images from RTX cameras and generates asynchronous event streams. These streams are published through ROS~2 to external applications running in a host environment managed with \texttt{pixi}, enabling seamless integration with visualization, recording, and robotics software while maintaining a clear separation between the simulation and application environments.

\subsection{Event Generation}
First, rendered images are converted into asynchronous events using a configurable log-intensity model. Subsequently, motion-guided frame-gap synthesis is employed to increase the effective temporal resolution without increasing the rendering frequency of Isaac Sim.

\subsubsection{Log-Intensity Event Model}
The default event generation model in \addon follows a frame-based log-intensity formulation. Given an HDR image $I_k$, it is first converted into a scalar intensity representation $S_k$ using grayscale, weighted luminance, or a Bayer Red-Green-Green-Blue (RGGB) sampling pattern, corresponding to the native color filter array used by Bayer image sensors. The logarithmic intensity is then computed as
\begin{equation}
L_k(x,y)=\log\left(S_k(x,y)+1\right),
\end{equation}
and compared against a per-pixel reference state
\begin{equation}
\Delta L_k(x,y)=L_k(x,y)-L_{\mathrm{ref}}(x,y).
\end{equation}
An ON or OFF event is generated whenever the accumulated contrast satisfies
\begin{equation}
\Delta L_k(x,y)\ge C_{\mathrm{ON}},
\qquad
\Delta L_k(x,y)\le -C_{\mathrm{OFF}},
\end{equation}
producing an event
$
e=(x,y,t,p).
$
The corresponding reference intensity is then updated before processing the next sample.

For Bayer RGGB event generation, the scalar intensity representation $S_k$ is obtained from the virtual RGGB mosaic rather than a grayscale conversion. Consequently, each pixel stores only one sampled color component (R, G$_1$, G$_2$, or B), and the same log-intensity formulation is applied independently to the corresponding channel. This approach preserves the spatial sampling pattern of Bayer-based event cameras while maintaining the same event generation model.

\subsubsection{Motion-Guided Frame-Gap Synthesis}
Although the event generation model operates on discrete rendered images, the temporal resolution can be increased through motion-guided frame-gap synthesis. Rather than increasing the rendering frequency, \addon synthesizes intermediate intensity images using the motion vectors provided by Isaac Sim and processes them using the same log-intensity event generation model. 

Given two consecutive rendered images $I_{k-1}$ and $I_k$, the renderer provides a dense motion vector field
$
\mathbf{v}(x,y)=\bigl(u(x,y),v(x,y)\bigr),
$
which describes the image-space displacement between consecutive rendered images. Intermediate samples are synthesized by backward warping the current rendered image according to a temporally scaled motion field. For an interpolation factor
$
\alpha=\frac{n}{N},
n=1,\ldots,N-1,
$
the synthesized intensity image is obtained as
\begin{equation}
I_{k,n}(x,y)
=
I_k\!\left(
x-\alpha u(x,y),
y-\alpha v(x,y)
\right),
\end{equation}
where bilinear interpolation is used for non-integer sampling locations. Each synthesized image is subsequently processed using the same log-intensity event generation model described above. If motion vectors are unavailable or invalid, the implementation falls back to linear image interpolation.

For $N$ temporal subdivisions between consecutive rendered images acquired at times $t_{k-1}$ and $t_k$, the timestamp of the $n$-th synthesized image is defined as
\begin{equation}
t_{k,n}
=
t_{k-1}
+
\frac{n}{N}
\left(
t_k-t_{k-1}
\right),
\qquad
n=1,\ldots,N-1.
\end{equation}

\begin{figure}[!h]
    \centering
    \resizebox{\columnwidth}{!}{%
    \begin{tikzpicture}[
        font=\footnotesize,
        >=Latex,
        rendered/.style={
            draw=black,
            rounded corners=2pt,
            minimum width=1.45cm,
            minimum height=0.65cm,
            align=center,
            fill=white
        },
        intermediate/.style={
            draw=black,
            rounded corners=2pt,
            minimum width=1.45cm,
            minimum height=0.65cm,
            align=center,
            fill=blue!14
        },
        process/.style={
            draw=black,
            rounded corners=2pt,
            minimum width=3.35cm,
            minimum height=0.68cm,
            align=center,
            fill=gray!10
        },
        output/.style={
            draw=black,
            rounded corners=2pt,
            minimum width=1.95cm,
            minimum height=0.68cm,
            align=center,
            fill=green!10
        },
        synthesis/.style={
            draw=black,
            rounded corners=2pt,
            minimum width=2.55cm,
            minimum height=0.60cm,
            align=center,
            fill=blue!7
        },
        temporal/.style={->, thick},
        flow/.style={->, thick},
        auxiliary/.style={->, thick, dashed}
    ]

    \node[rendered] (prev) {
        \scriptsize $I_{k-1}$\\[-1pt]
        \scriptsize $t_{k-1}$
    };

    \node[intermediate, right=0.45cm of prev] (s1) {
        \scriptsize $I_{k,1}$\\[-1pt]
        \scriptsize $t_{k,1}$
    };

    \node[intermediate, right=0.40cm of s1] (s2) {
        \scriptsize $I_{k,2}$\\[-1pt]
        \scriptsize $t_{k,2}$
    };

    \node[right=0.30cm of s2] (dots) {$\cdots$};

    \node[intermediate, right=0.30cm of dots] (sn) {
        \scriptsize $I_{k,N-1}$\\[-1pt]
        \scriptsize $t_{k,N-1}$
    };

    \node[rendered, right=0.45cm of sn] (curr) {
        \scriptsize $I_k$\\[-1pt]
        \scriptsize $t_k$
    };

    \draw[temporal] (prev) -- (s1);
    \draw[temporal] (s1) -- (s2);
    \draw[temporal] (s2) -- (dots);
    \draw[temporal] (dots) -- (sn);
    \draw[temporal] (sn) -- (curr);

    \node[
        synthesis,
        above=0.55cm of dots
    ] (synthesis) {
        \scriptsize Motion-guided\\
        \scriptsize interpolation
    };

    \draw[auxiliary]
        (prev.north)
        to[out=65,in=180]
        (synthesis.west);

    \draw[auxiliary]
        (curr.north)
        to[out=115,in=0]
        (synthesis.east);

    \draw[auxiliary]
        (synthesis.south)
        --
        ($(dots.north)+(0,0.10)$);

    \node[
        process,
        below=1.15cm of dots
    ] (backend) {
        \scriptsize Event generation model
    };

    \node[
        output,
        right=0.60cm of backend
    ] (events) {
        \scriptsize Asynchronous events\\[-1pt]
        \scriptsize $(x,y,t,p)$
    };

    \coordinate (busleft) at ($(prev.south)+(0,-0.45)$);
    \coordinate (busright) at ($(curr.south)+(0,-0.45)$);

    \draw[thick] (busleft) -- (busright);

    \foreach \sample in {prev,s1,s2,sn,curr}{
        \draw[flow]
            (\sample.south)
            --
            (\sample.south |- busleft);
    }

    \draw[flow]
        ($(backend.north |- busleft)$)
        --
        (backend.north);

    \draw[flow] (backend) -- (events);

    \end{tikzpicture}%
    }
    \caption{
        Motion-vector-based frame-gap synthesis in \addon.
        Given two consecutive rendered images, motion information is used to
        synthesize $N-1$ intermediate intensity samples at uniformly spaced
        timestamps. Both rendered and synthesized samples are processed by the
        same frame-based event backend, producing a unified asynchronous event
        stream.
    }
    \label{fig:frame-gap-synthesis}
\end{figure}
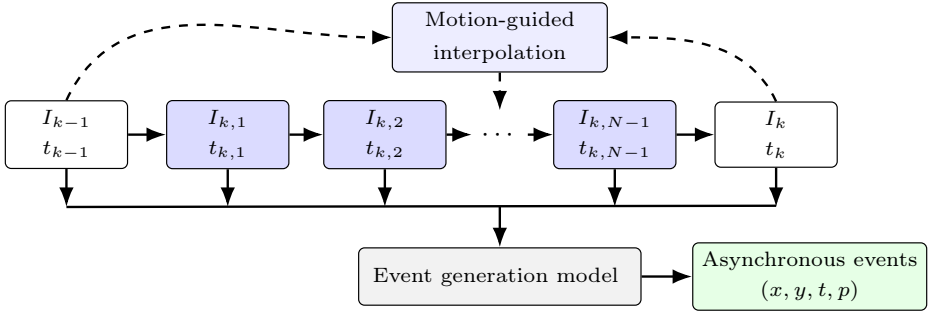

Figure~\ref{fig:frame-gap-synthesis} illustrates the motion-guided frame synthesis pipeline. Given two consecutive rendered images and their associated motion vectors, intermediate intensity images are synthesized at uniformly spaced timestamps. Both rendered and synthesized images are subsequently processed by the same event generation stage, producing a unified asynchronous event stream.

\subsection{Sensor Model Configuration}
\addon exposes a configurable sensor model that allows users to adapt the event generation process to different event camera characteristics without modifying the underlying implementation. Sensor parameters are specified at runtime and independently configured for each simulated camera, enabling multiple virtual sensors with different characteristics to coexist within the same Isaac Sim scene.

The current implementation supports configurable contrast thresholds, threshold mismatch, refractory period, latency, timestamp quantization, shot and leak noise, and grayscale or Bayer RGGB intensity generation. These parameters affect only the event generation backend while remaining independent of the rendering pipeline, allowing the same rendered images to be reused under different sensor configurations.

\begin{table}[!h]
\centering
\caption{Configurable parameters of the \addon sensor model. The simulator exposes runtime-configurable settings controlling spatial resolution, event generation thresholds, temporal behavior, noise characteristics, and intensity conversion modes.}
\label{tab:sensor_parameters}
\small
\begin{tabular}{@{}ll@{}}
\toprule
\textbf{Parameter} & \textbf{Description} \\
\midrule
Spatial resolution & $128\times128$, $240\times180$, $304\times240$, $260\times346$, $320\times320$ \\
$C_{\mathrm{ON}}$, $C_{\mathrm{OFF}}$ & Positive and negative contrast thresholds \\
Threshold mismatch & Pixel-wise threshold variability \\
Refractory period & Minimum time between consecutive events \\
Latency & Event output delay \\
Timestamp quantization & Timestamp discretization resolution \\
Shot noise & Random spurious events \\
Leak noise & Background activity generation \\
Intensity mode & Grayscale, luminance, or Bayer RGGB \\
\bottomrule
\end{tabular}
\end{table}

\subsection{Output Interfaces}
\begin{figure}[!h]
\centering
    \includegraphics[width=0.8\linewidth]{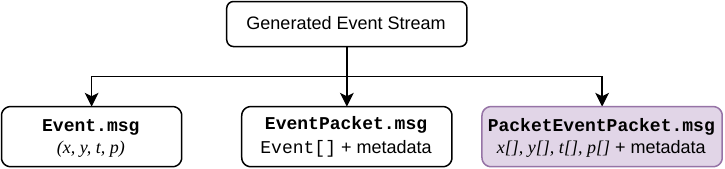}
    \caption{ROS~2 message interfaces provided by \addon for event-stream communication. Individual events are represented by \texttt{Event}, conventional batched communication uses \texttt{EventPacket}, and \texttt{PackedEventPacket} adopts a structure-of-arrays layout to reduce serialization overhead in high-rate event streams.}
    \label{fig:output_interfaces}
\end{figure}

To facilitate integration with robotics applications, \addon provides dedicated ROS~2 message interfaces for event streams. As illustrated in Fig.~\ref{fig:output_interfaces}, generated events can be represented individually, grouped into conventional event packets, or encoded using a packed Structure-of-Arrays layout for high-rate communication.

In addition to event streams, \addon can publish synchronized auxiliary data generated by Isaac Sim, including APS intensity images (grayscale or Bayer RGGB), depth images, and IMU measurements. These data streams are timestamped consistently with the generated events, enabling multimodal perception pipelines and facilitating the development and evaluation of event-based visual-inertial, RGB-event, and depth-event algorithms.

\section{Results and Discussion}
\addon was evaluated through qualitative and performance analyses. First, representative simulation outputs are presented to demonstrate the framework's capability to generate synchronized multimodal data, including RGB, APS grayscale, event, and depth streams. Subsequently, the computational performance of the event generation pipeline is analyzed across multiple sensor resolutions and temporal interpolation factors.

\subsection{Qualitative Evaluation}
\begin{figure}[!h]
    \centering
    \includegraphics[width=\columnwidth]{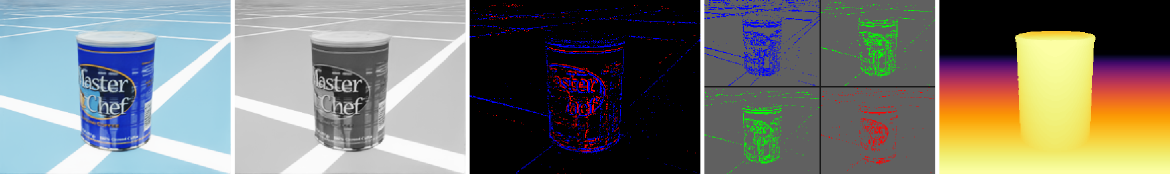}
    \includegraphics[width=\columnwidth]{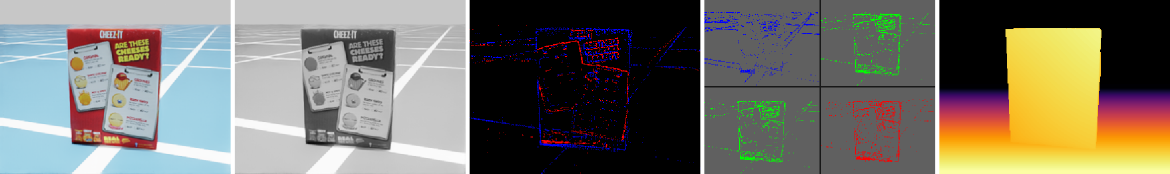}
    \includegraphics[width=\columnwidth]{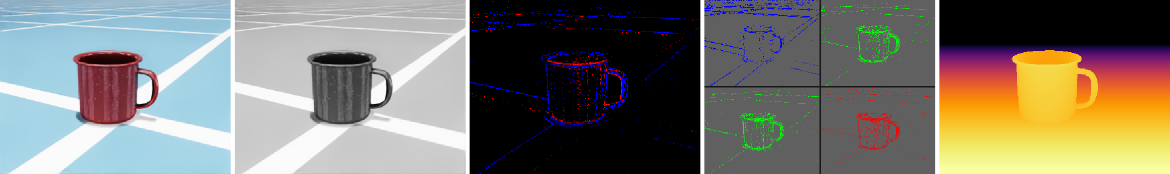}
    \includegraphics[width=\columnwidth]{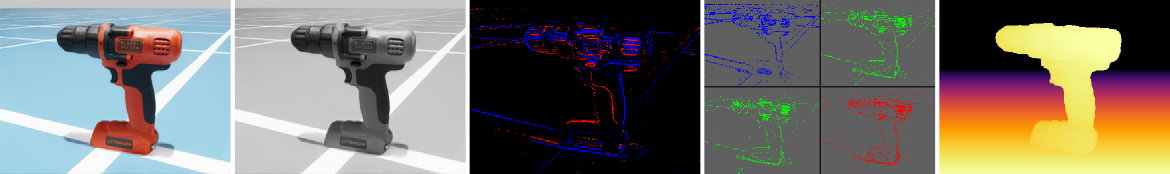}
    \includegraphics[width=\columnwidth]{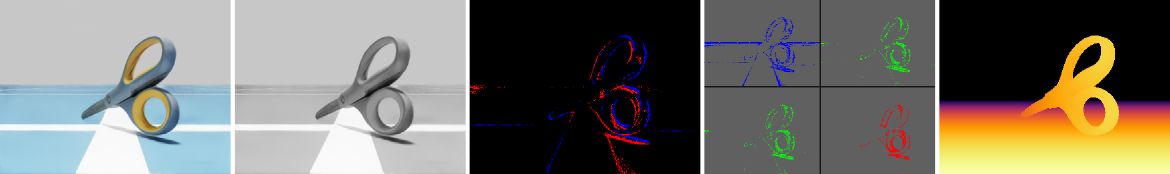}
    \caption{
    Qualitative results produced by \addon on representative YCB objects using a simulated DAVIS346 event camera ($346\times260$ resolution). Each row corresponds to a different object, while columns show the synchronized RGB image, APS grayscale image, grayscale event visualization, Bayer RGGB event visualization, and depth image generated from the same simulated scene.
    }
    \label{fig:qualitative_results}
\end{figure}

Figure~\ref{fig:qualitative_results} presents representative outputs generated by \addon for a subset of YCB objects~\cite{calli2015ycb} with diverse geometries, textures, and material properties. For each object, the simulator simultaneously produces an RGB image, an APS grayscale image, a grayscale event visualization, a Bayer RGGB event visualization, and a depth image from the same virtual sensor configuration.

The results show that \addon consistently generates synchronized multimodal outputs while preserving object appearance and geometric structure across different sensing modalities. The grayscale and Bayer event visualizations capture object contours and intensity transitions induced by camera motion, whereas the depth images accurately represent the three-dimensional scene geometry. Together, these examples illustrate the flexibility of the proposed framework for simulating heterogeneous event-camera data for photorealistic robotic environment applications.

\subsection{Sensor Model Evaluation}

\begin{figure}[!h]
    \centering

    \begin{minipage}[c]{0.45\columnwidth}
        \centering
        \includegraphics[width=\linewidth]{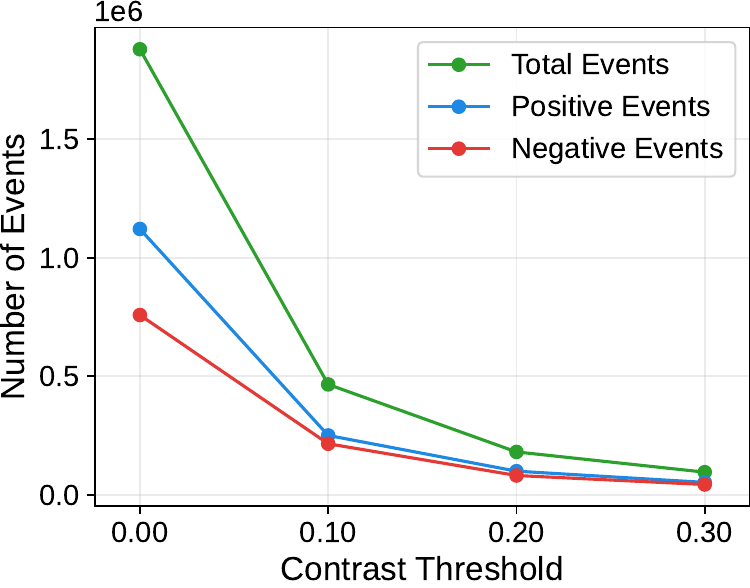}
    \end{minipage}
    \quad
    \begin{minipage}[c]{0.43\columnwidth}
        \centering
        \includegraphics[width=.48\linewidth]{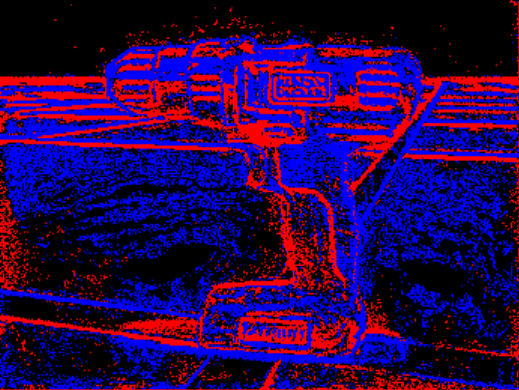}
        \hfill
        \includegraphics[width=.48\linewidth]{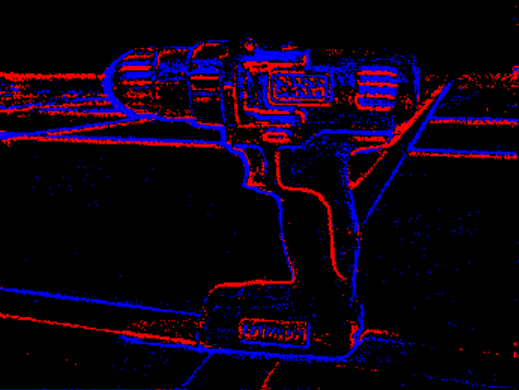}

        \vspace{1mm}

        \includegraphics[width=.48\linewidth]{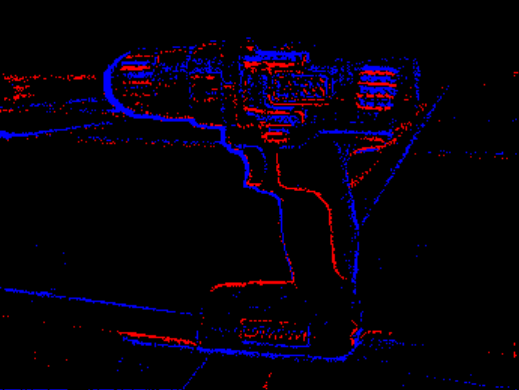}
        \hfill
        \includegraphics[width=.48\linewidth]{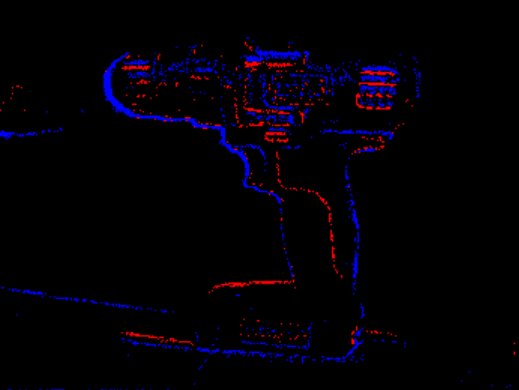}
    \end{minipage}

    \caption{
    Effect of the configurable contrast threshold on event generation. Left: total, positive, and negative event counts obtained from the same Power Drill sequence while varying the contrast threshold. Right: corresponding event visualizations for thresholds of 0.00, 0.10, 0.20, and 0.30. Increasing the threshold reduces the generated event density while preserving the dominant scene structure.
    }
    \label{fig:sensor_parameter_sweep}
\end{figure}

Figure~\ref{fig:sensor_parameter_sweep} illustrates the influence of the configurable sensor model by varying the contrast threshold while maintaining the same Power Drill sequence, camera trajectory, and illumination conditions. 

As expected from the event generation model, increasing the contrast threshold substantially reduces the number of generated events because larger logarithmic intensity variations are required to trigger events. Despite the reduced event density, the dominant object contours and motion-induced structures remain visible across all configurations. These results show that the proposed framework enables runtime-configurable event camera behavior while preserving the qualitative characteristics of the generated event stream.

\subsection{Performance Evaluation}
To assess the computational efficiency of \addon, the event generation pipeline was benchmarked on a local workstation equipped with an NVIDIA GeForce RTX~4060 GPU with 8~GB of GPU memory. This consumer-grade setup was used without access to high-performance computing infrastructure. Five representative event camera resolutions (DVS128, DAVIS240, ATIS, DAVIS346, and GenX320) were evaluated under temporal interpolation factors of $\times8$, $\times16$, and $\times32$. Assuming a 30~Hz rendering cadence, these factors correspond to effective temporal sampling frequencies of 240, 480, and 960~Hz, respectively. The evaluation measures only the event generation pipeline, excluding scene loading, rendering initialization, ROS~2 communication, logging, and file I/O.

\begin{figure}[!h]
    \centering
    \includegraphics[width=\columnwidth]{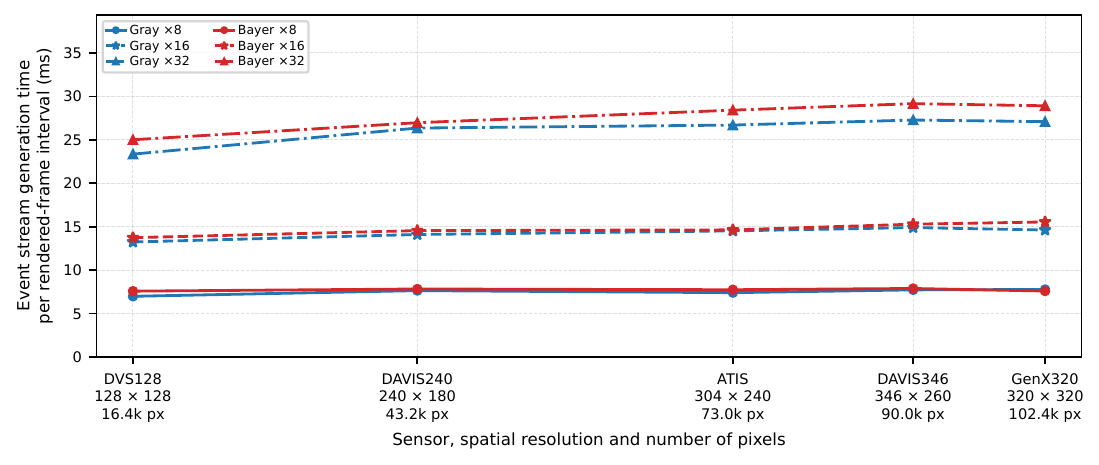}
    \caption{
    Event stream generation time per rendered-frame interval across sensor resolutions for grayscale and Bayer RGGB event generation under temporal interpolation factors of $\times8$, $\times16$, and $\times32$. Assuming a 30~Hz rendering cadence, these correspond to effective temporal sampling frequencies of 240, 480, and 960~Hz, respectively.
    }
    \label{fig:event_generation_time}
\end{figure}

Figure~\ref{fig:event_generation_time} shows the average event stream generation time per rendered-frame interval as a function of sensor resolution for both grayscale and Bayer RGGB event generation. The reported execution times correspond to the processing time required by the event-generation backend to generate the complete event stream associated with one interval between two consecutive rendered frames. Therefore, the effective temporal sampling frequencies characterize the temporal resolution of the generated event stream rather than the processing throughput of the simulator.

As expected, the computational cost generally increases with both sensor resolution and temporal interpolation factor because additional synthesized intermediate samples must be processed within each rendered-frame interval. Across the five evaluated sensor resolutions, event stream generation ranged from 6.98 to 27.28~ms for grayscale mode and from 7.58 to 29.16~ms for Bayer RGGB mode. Small non-monotonic variations are observed between sensors with similar pixel counts due to implementation and memory-access characteristics, while the overall trend remains consistent.

Table~\ref{tab:gpu_memory} summarizes the GPU memory allocated during grayscale and Bayer RGGB event generation. For both modes, memory consumption scales approximately linearly with sensor resolution and temporal interpolation factor, ranging from 51.4~MB for DVS128 at $\times8$ to 399.7~MB for GenX320 at $\times32$. Bayer RGGB event generation incurs only a marginal increase in GPU memory, with differences of at most 1.5~MB compared with grayscale mode.

\begin{table}[!t]
\centering
\caption{
GPU memory (MB) allocated during event generation for grayscale and Bayer RGGB modes. Time interpolation factors $\times8$, $\times16$, and $\times32$ correspond to effective event rates of 240, 480, and 960 Hz, respectively.
}
\label{tab:gpu_memory}

\small
\setlength{\tabcolsep}{4pt}
\renewcommand{\arraystretch}{1.05}

\begin{tabular}{@{}lccccccc@{}}
\toprule
\multirow{2}{*}{\textbf{Sensor}} &
\multirow{2}{*}{\textbf{Resolution}} &
\multicolumn{2}{c}{$\times8$} &
\multicolumn{2}{c}{$\times16$} &
\multicolumn{2}{c}{$\times32$} \\
\cmidrule(lr){3-4}
\cmidrule(lr){5-6}
\cmidrule(lr){7-8}
&
&
\textbf{Gray} &
\textbf{Bayer} &
\textbf{Gray} &
\textbf{Bayer} &
\textbf{Gray} &
\textbf{Bayer} \\
\midrule
DVS128   & 128$\times$128 &  51.4 &  51.4 &  55.4 &  55.4 &  64.0 &  64.1 \\
DAVIS240 & 240$\times$180 & 135.5 & 135.6 & 146.0 & 146.2 & 167.8 & 168.0 \\
ATIS     & 304$\times$240 & 228.7 & 229.0 & 246.5 & 246.8 & 282.4 & 282.7 \\
DAVIS346 & 346$\times$260 & 287.0 & 288.5 & 310.4 & 310.8 & 354.3 & 354.6 \\
GenX320  & 320$\times$320 & 324.2 & 324.6 & 349.2 & 349.7 & 399.3 & 399.7 \\
\bottomrule
\end{tabular}
\end{table}

Even under the most demanding evaluated configuration, the event generation backend required less than 400~MB of additional GPU memory, leaving substantial capacity available for the Isaac Sim rendering pipeline and other robotic perception components on the evaluated 8~GB GPU.
Overall, these results indicate that \addon efficiently generates both gray and Bayer RGGB event streams across a range of commonly used event camera resolutions while maintaining predictable computational and memory scaling. Importantly, all experiments were conducted on a local consumer-grade RTX~4060 GPU rather than specialized HPC hardware. The measured execution times and moderate memory requirements therefore suggest that \addon can support online event simulation on an accessible workstation, facilitating development and experimentation even when dedicated high-performance computing resources are unavailable.

Despite these promising results, several limitations remain in the current version of \addon. First, the event generation backend is currently frame-based only; the physics-grounded backend reserved in the sensor interface has not yet been implemented, so effects such as optical path modeling and pixel-level circuit behavior are not yet captured. Second, the reported evaluation focuses on computational efficiency and qualitative multimodal consistency, without a quantitative comparison against real event camera recordings; establishing this correspondence is necessary to assess the fidelity of the generated event streams. Third, all experiments were conducted on a single consumer-grade GPU with a limited set of static and quasi-static robotic scenes, and further evaluation under more dynamic manipulation tasks and multi-camera configurations is needed to characterize performance under more demanding robotic workloads.

\section{Conclusions and Future Work}

This paper presented \addon, a multimodal event camera add-on for NVIDIA Isaac Sim that enables online simulation of asynchronous event streams within photorealistic robotic environments. By integrating directly into the native Isaac Sim rendering pipeline, the proposed framework provides synchronized RGB, APS grayscale, grayscale event, Bayer RGGB event, IMU, and depth outputs while remaining fully compatible with existing Isaac Sim and ROS~2 workflows.

Experimental results showed efficient online event simulation across five representative event camera resolutions. Event streams corresponding to a single rendered-frame interval were generated in 6.98--27.28~ms for grayscale mode and 7.58--29.16~ms for Bayer RGGB mode while requiring less than 400~MB of additional GPU memory on a consumer-grade NVIDIA RTX~4060 GPU. Motion-guided temporal interpolation enabled effective temporal sampling rates of up to 960 Hz while maintaining online processing of each rendered-frame interval.

The current version of \addon nonetheless presents several limitations that motivate ongoing development. The event generation pipeline remains frame-based, with the physics-grounded backend interface defined but not yet active, limiting the sensor-level realism achievable relative to native circuit-level event models. In addition, the generated event streams have not yet been quantitatively validated against real event camera data, and the evaluation reported in this work does not yet include downstream robotic perception tasks.

Future work will focus on extending the sensor model with additional event camera effects, including more realistic noise sources and hardware-specific characteristics, as well as incorporating support for additional event sensor configurations. We also plan to implement and evaluate the physics-based backend, validate the simulated event streams against real event camera data, and assess their effectiveness in downstream robotic perception tasks such as visual odometry, SLAM, object detection, and event-based learning.

\section*{Acknowledgements}

This research was supported by the NVIDIA Academic Grant Program 2025, Robotics and Edge AI Track.
%
%

\bibliographystyle{splncs04}
\bibliography{main}
\end{document}